%% file: colm2024_conference.tex
\documentclass[dvipsnames]{article} %
\usepackage{colm2024_conference}

\usepackage{booktabs}
\usepackage{graphicx}
\usepackage{enumitem}
\usepackage{wrapfig}
\usepackage{algorithm}
\usepackage{algpseudocode}

\usepackage{microtype}
\usepackage{amsmath}
\usepackage{colortbl}
\usepackage[utf8]{inputenc}
\definecolor{lightgray}{rgb}{0.9,0.9,0.9}
\usepackage{caption}
\usepackage{subcaption}
\usepackage[dvipsnames]{xcolor}
\usepackage{setspace}
\usepackage{url}
\usepackage{multirow}
\usepackage{colortbl}
\usepackage{tabularx}
\usepackage{blindtext}
\usepackage{pgfplots}
\pgfplotsset{compat=1.18} 
\usepackage{tikz}
\usetikzlibrary{er,positioning,bayesnet}
\usepackage{makecell}
\usepackage{tipa}
\usepackage{siunitx}
\usepackage{nicefrac}
\usepackage{tocloft}
\usepackage{listings}
\usepackage[raster,skins]{tcolorbox} %
\usepackage{xltabular}
\usepackage{adjustbox}
\usepackage{xurl}
\usepackage{rotating}
\usepackage[normalem]{ulem}
\usepackage{booktabs}      
\usepackage{multirow}      
\usepackage{threeparttable}
\usepackage{makecell}      
\usepackage{array}
\usepackage{amsmath}
\usepackage{wrapfig}

\usepackage{colortbl}

\usepackage{pifont}        
\usepackage{amssymb}       
\usepackage{wasysym}       

\usepackage{xcolor}

\input{math_commands.tex}

\renewcommand{\ghlink}{https://github.com/JhCircle/Kardia-R1}

\newcommand*\justify{%
  \fontdimen2\font=0.4em
  \fontdimen3\font=0.2em
  \fontdimen4\font=0.1em
  \fontdimen7\font=0.1em
  \hyphenchar\font=`\-
}

\renewcommand{\texttt}[1]{%
  \begingroup
  \ttfamily
  \begingroup\lccode`~=`/\lowercase{\endgroup\def~}{/\discretionary{}{}{}}%
  \begingroup\lccode`~=`[\lowercase{\endgroup\def~}{[\discretionary{}{}{}}%
  \begingroup\lccode`~=`.\lowercase{\endgroup\def~}{.\discretionary{}{}{}}%
  \catcode`/=\active\catcode`[=\active\catcode`.=\active
  \justify\scantokens{#1\noexpand}%
  \endgroup
}

\title{Kardia-R1: Unleashing LLMs to Reason toward Understanding and Empathy for Emotional Support via Rubric-as-Judge Reinforcement Learning}

\author{
\bf Jiahao Yuan\thanks{Project Leader: \url{jhyuan.cs@gmail.com}} \quad \bf Zhiqing Cui \quad  \bf Hanqing Wang \quad Yuansheng Gao \\
\bf Yucheng Zhou\thanks{Corresponding Author: \url{yucheng.zhou@connect.um.edu.mo}} \quad \bf Usman Naseem \\
}

\begin{document}

\maketitle

\begin{abstract}
As web platforms evolve towards greater personalization and emotional complexity, conversational agents must transcend superficial empathy to demonstrate identity-aware emotional reasoning. 
However, existing systems face two limitations: (1) reliance on situation-centric datasets lacking persistent user identity, which hampers the capture of personalized affective nuances; and (2) dependence on opaque, coarse reward signals that hinder development of verifiable empathetic reasoning. 
To address these gaps, we introduce KardiaBench, a large-scale user-grounded benchmark comprising 178,080 QA pairs across 22,080 multi-turn conversations anchored to 671 real-world profiles. 
The dataset is constructed via a model-in-the-loop pipeline with iterative rubric-guided refinement to ensure psychological plausibility and persona consistency. 
This progressive empathy pipeline that integrates user comprehension, contextual reasoning, and emotion perception into conversations, followed by iterative critique and rubric-based refinement to ensure psychological plausibility, emotional fidelity, and persona consistency.
Building on this, we propose Kardia-R1, a framework that trains models for interpretable, stepwise empathetic cognition. 
Kardia-R1 leverages \textbf{Rubric-as-Judge Empathetic Reinforcement Learning (Rubric-ERL)}, a GRPO-based method that uses explainable, human-aligned rubric rewards to tightly couple user understanding, emotional inference, and supportive response generation. 
Extensive experiments across four LLM backbones demonstrate that Kardia-R1 consistently outperforms othet methods in emotion accuracy, empathy, relevance, persona consistency, and safety. Our dataset and model will be released at \url{https://github.com/JhCircle/Kardia-R1}.

\end{abstract}


\input{content/intro.tex}
\input{content/relatedwork}
\input{content/method}

\input{content/experiments.tex}
\input{content/conclusion.tex}
\input{content/authors.tex}

\bibliography{biblio}
\bibliographystyle{colm2024_conference}
\appendix
\input{content/appendix.tex}

\end{document}

%% file: math_commands.tex
\usepackage{amsmath,amsfonts,bm}

\def\eqref#1{equation~\ref{#1}}

\def\1{\bm{1}}

\DeclareMathAlphabet{\mathsfit}{\encodingdefault}{\sfdefault}{m}{sl}
\SetMathAlphabet{\mathsfit}{bold}{\encodingdefault}{\sfdefault}{bx}{n}



%% file: content/intro.tex
\begin{figure*}[t]
    \centering
    \includegraphics[width=0.66\linewidth]{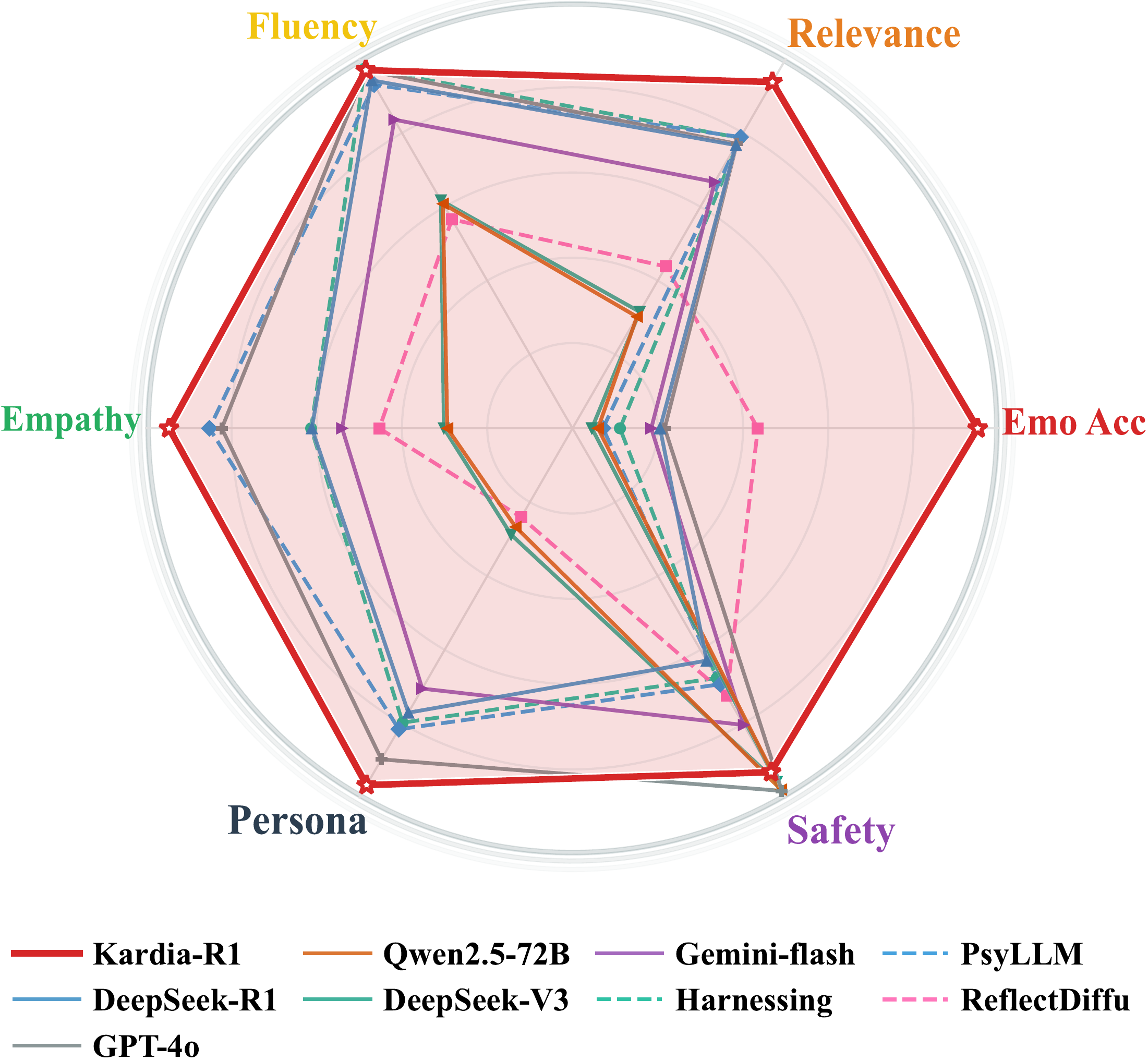}
    \caption{Kardia-R1 versus strong general-purpose LLMs and specialized empathetic systems across six core dimensions of empathetic dialogue (emotion recognition accuracy, relevance, fluency, safety, persona consistency, and empathy).}
    \label{fig:intro-radar}
\end{figure*}

\section{Introduction}
\label{sec:intro}

As web-based conversational AI \cite{zhang2020dialogpt,yang2024mentallama} becomes increasingly integrated into daily life, online platforms have evolved into ``emotional commons'' where individuals share vulnerabilities and complex affective states \cite{miller2011social}. Supporting such interactions requires models that provide user-grounded emotional support shaped by personal background and situational needs. This depends on systems able to perceive emotion, reason about context, and account for individual differences. Empathy, central to theories of trust and perspective taking \cite{batson1991empathic}, is therefore essential for meaningful human–AI interaction. Although recent work adapts to emotional cues \cite{gao2021improving,gao2023cab} or models affective states \cite{majumder2020mime,bi2023diffusemp,yuan2025reflectdiffu}, current systems still struggle to incorporate users’ identities and nuanced emotional histories.

Existing empathetic dialogue models can be broadly categorized into two categories: small-scale models that integrate explicit empathy mechanisms, and LLM-based models that leverage reasoning capabilities \cite{dubey2024llama,team2024qwen2}. Small-scale models enhance contextual understanding through emotion-conditioned decoding \cite{lin2019moel,majumder2020mime}, multi-resolution knowledge \cite{li2020empdg,li2022knowledge}, or multi-grained signals including emotional cause \cite{bi2023diffusemp,hamad2024asem} and emotion-intent reflection \cite{yuan2025reflectdiffu}, enabling emotionally responsive replies. However, their reliance on heuristic signals and limited external knowledge often yields fluent but shallow responses that misalign with nuanced user affect. In contrast, LLM-based models exploit instruction tuning \cite{chen2023soulchat} and chain-of-thought reasoning \cite{chen2023soulchat,hu2024aptness,cai2024empcrl} to simulate higher-order empathetic cognition via multi-step inference. 

Building on datasets like EmpatheticDialogues \cite{rashkin2019towards}, recent work further leverages LLMs to synthesize auxiliary supervision, such as emotional causes \cite{he2025ecc}, user intent \cite{xie2021empathetic,yuan2025reflectdiffu} and personality traits \cite{wu2025traits} for finetuning \cite{bi2023diffusemp,chen2025synthempathy} or reinforcement learning \cite{li2024reinforcement,dai2025psyche}. However, these methods or benchmarks remain detached from users' concrete backgrounds and emotional states, overlooking perspective-taking principles \cite{batson1991empathic} that stress grounding empathy in an individual's specific context and experiences. In summary, current LLM-based empathetic dialogue models face two major challenges: \textbf{(1) lack of user-grounded training data}, as most benchmarks, including EmpatheticDialogues, represent general emotional situations rather than user-oriented interactions, limiting models’ ability to generate responses aligned with individual user's nuanced emotional states; \textbf{(2) challenges in training with verifiable reward signals}, since supervised fine-tuning and reinforcement learning with reward models are difficult to validate in empathetic dialogue, as existing evaluation metrics cannot reliably assess whether responses genuinely reflect users' affective states and contextual subtleties.

To address these limitations, we first construct \textbf{KardiaBench}, a user-grounded multi-turn empathetic dialogue benchmark derived from 671 real online profiles and EmpatheticDialogues \cite{rashkin2019towards}. In our pipeline, one LLM simulates a user generating queries based on a specific profile and situation, while another acts as an empathetic expert performing perspective-taking. Crucially, each turn is iteratively refined via user critique and model-in-the-loop updates under rubric evaluation, and finally verified through manual checks. Building on this benchmark, we develop \textbf{Kardia-R1}. This model is trained via supervised fine-tuning (SFT) for cold-start alignment, followed by GRPO-based reinforcement learning \cite{guo2025deepseek} guided by \textbf{Rubric-as-Judge Empathetic RL (Rubric-ERL)}. This framework enables the model to actively learn user comprehension and empathetic generation, using turn-level critique to optimize its policy for coherence and alignment with user background, personality, and emotional state. 

Our main contributions are summarized as follows:
\begin{itemize}[left=0pt]
\item We propose \textbf{KardiaBench}, a user-grounded multi-turn empathetic dialogue benchmark that augments EmpatheticDialogues with 671 real online user profiles, leveraging a fine-grained LLM data synthesis pipeline to generate high-quality multi-turn empathetic dialogues. 
\item We develop \textbf{Kardia-R1}, a progressive framework that models empathy as a perspective-taking process from comprehension and reasoning to empathetic response generation, trained via finetuning and GRPO-based reinforcement learning guided by verifiable rubric-based rewards.
\item Extensive experiments show that Kardia-R1 outperforms reward-model methods, embedding-based rubrics, and strong empathetic-dialogue baselines, achieving scalable state-of-the-art performance in automatic and human evaluation.
\end{itemize}

%% file: content/relatedwork.tex
\section{Related Work}

\subsection{Empathetic Dialogue Generation}

Empathetic dialogue generation aims to produce contextually aligned responses that track users’ emotions and needs rather than merely echoing surface sentiment \cite{batson1991empathic}. Early work largely relies on small-scale language models equipped with explicit empathy mechanisms, employing emotion-conditioned or mimicking decoding based on emotion-specific experts or polarity-aware mixtures~\cite{lin2019moel,majumder2020mime}, incorporating multi-resolution emotion modeling and external affective knowledge to bridge dialogue-level and token-level cues  \cite{bi2023diffusemp,hamad2024asem}, and injecting multi-grained affective signals—such as emotional causes, intents, and their interactions—to strengthen alignment between user states and generated responses~\cite{yuan2025reflectdiffu}. While these methods improve local affect sensitivity in controlled benchmarks, they remain constrained by heuristic labels, limited external knowledge, and narrow decoding heuristics, often producing fluent yet template-like responses that fail to capture subtle affective shifts or adapt to users’ idiosyncratic backgrounds.

Large language models shift empathetic dialogue toward a reasoning-centered paradigm \cite{wei2022chain,guo2025deepseek}, in which instruction-tuned assistants use counseling-style multi-turn data and structured prompts to analyze user experiences before generating supportive responses \cite{chen2023soulchat}. Prompt-engineering approaches further embed appraisal-theoretic or commonsense-driven reasoning \cite{hu2024aptness,dai2025psyche,cai2024empcrl}, encouraging models to infer emotional drivers, evaluate situational appraisals, articulate multi-step empathetic cognition, and calibrate emotional strategies beyond generic comfort. Additional methods synthesize affective signals—such as emotional causes, user intent, and personality traits \cite{he2025ecc,xie2021empathetic,yuan2025reflectdiffu,wu2025traits}—and incorporate them into finetuning \cite{bi2023diffusemp} or reinforcement-learning pipelines \cite{li2024reinforcement,sotolar2024empo,dai2025psyche} via embedding-based rewards on EmpatheticDialogues \cite{rashkin2019towards}. Although such LLM-based emotional support agents demonstrate strong cognitive-empathy reasoning and stable affective language \cite{xu2025multiagentesc}, they still struggle to sustain user-specific alignment and safe long-horizon behavior as needs evolve, and they incur high inference and adaptation costs when required to cover increasingly diverse emotional scenarios. We address these inefficiencies by training LLMs to internalize a perspective-taking empathy that integrates user comprehension, contextual reasoning, emotion recognition, and empathetic response generation, thereby improving generalization to new users and shifting emotional contexts.

\subsection{Benchmarking for Empathetic Dialogue}
\begin{table*}[!htp]\small
\centering
\caption{
Comparison of empathetic dialogue benchmarks across interaction depth, user grounding, reasoning supervision, and empathy modeling. 
\Circle: feature absent; 
\LEFTcircle: partially supported (e.g., shallow traits or limited reasoning labels); 
\CIRCLE: fully supported with explicit, structured supervision or profile-level grounding.
}
\resizebox{\textwidth}{!}{
\begin{tabular}{lcccccccc}
\toprule
\textbf{Dataset} 
& \textbf{\makecell{Avg.\\Turns}} 
& \textbf{\makecell{\#Convs\\(Train / Test)}} 
& \textbf{\makecell{\#Pairs\\(Train / Test)}} 
& \textbf{Source} 
& \textbf{User}
& \textbf{Underst.}
& \textbf{Reason.} 
& \textbf{Emp.} \\
\midrule

\textsc{ED \cite{rashkin2019towards}}
& 2.06 
& 19.5k+ / 2.5k+
& 40.2k+ / 5.2k+ 
& Crowd
& $\emptyset$
& \Circle 
& \Circle
& \CIRCLE \\

\textsc{ESConv \cite{liu2021towards}} 
& 14.5
& 0.9k+ / 0.02k+
& 13.1k+ / 2.9k+
& Crowd
& $\emptyset$ 
& \Circle 
& \Circle
& \CIRCLE \\

\textsc{SODA~\cite{kim2023soda}} 
& 3.63
& 4.3M+ / 531k+ 
& 1.19M+ / 146k 
& LLMs
& $\emptyset$
& \Circle 
& \LEFTcircle
& \CIRCLE \\

\textsc{Big5-Chat~\cite{li2025big5}} 
& 1
& 100k / $\emptyset$ 
& 100k / $\emptyset$  
& LLMs
& LLMs
& \LEFTcircle
& \LEFTcircle
& \Circle \\

\textsc{ECC~\cite{he2025ecc}} 
& 2 
& 1.9k / 0.02k+ 
& 4.8k / 1.2k 
& LLMs 
& $\emptyset$
& \Circle 
& \LEFTcircle
& \CIRCLE \\

\midrule
\rowcolor{gray!10}
\textbf{\textsc{KardiaBench}}
& \textbf{8.07} 
& \textbf{19.5k+ / 2.5k+}
& \textbf{157k+ / 20k+} 
& \textbf{LLMs}
& \textbf{\makecell{671 Users}}
& \CIRCLE 
& \CIRCLE
& \CIRCLE \\
\bottomrule[1.5pt]
\end{tabular}
}
\label{tab:empathetic-benchmark-updated}
\end{table*}
Empathetic dialogue research has traditionally centered around situation-driven benchmarks such as \textsc{EmpatheticDialogues} (ED), which provides multi-turn conversations grounded in predefined emotional situations. While foundational, ED lacks persistent user profiles, emotional progression, and explicit reasoning supervision, limiting its ability to support evaluations of personalized and cognitively grounded empathy. Subsequent datasets such as \textsc{ESConv} \cite{liu2021towards} extend dialogue length and introduce counseling dynamics, yet remain confined to scenario-level interactions without modeling long-term user identity. Datasets like \textsc{ECC} \cite{he2025ecc} enrich ED with reasoning labels including emotion causes, intents, and outcomes, improving interpretability but maintaining a templated, situation-centric framework with limited scale. Large-scale LLM-generated corpora such as \textsc{SODA} \cite{kim2023soda} and \textsc{Big5-Chat} \cite{li2025big5} expand coverage and personality variation, but consist mostly of synthetic one-off dialogues loosely tied to user traits. As summarized in Table~\ref{tab:empathetic-benchmark-updated}, existing benchmarks offer partial support for reasoning and empathy, but lack realistic, user-grounded evaluation settings. Motivated by psychological theories of perspective-taking and individual-centered empathy~\cite{batson1991empathic}, we introduce \textsc{KardiaBench}, a user-grounded empathetic dialogue benchmark anchored in 671 real-world online profiles. It frames empathy as a progressive, profile-aware reasoning process across 22,000 multi-turn conversations. Each dialogue is constructed via an LLM-in-the-loop pipeline involving profile grounding, emotional triggering, user simulation, empathetic response generation, and rubric-guided revision. This process yields dialogues with stable user grounding, evolving emotional context, and interpretable reasoning chains. By aligning interactions with real identities and structured affective cues, \textsc{KardiaBench} enables more faithful evaluation of models' ability to understand, reason about, and support individual users, advancing empathy benchmarking beyond situation-centric or synthetic paradigms.

\subsection{Reinforcement Learning for Empathetic Cognition}

Supervised finetuning for empathy \cite{chen2025synthempathy} provides stylistic alignment but is limited in its ability to instill adaptive empathetic reasoning. Reinforcement learning (RL) offers an avenue for iterative policy refinement, yet existing methods often optimize reward models trained on implicit preferences or embedding similarities~\cite{li2024reinforcement}, offering limited transparency into reasoning quality or user alignment. For example, EmpCRL~\cite{cai2024empcrl} leverages commonsense priors to guide emotional tone, while Empo~\cite{sotolar2024empo} scores affective fit through sentence-level similarity—both lacking explicit grounding in user understanding or interpretable supervision. To bridge this gap, we propose Rubric-as-Judge Empathetic Reinforcement Learning (Rubric-ERL), a training paradigm that combines gradient-based RL with discrete, human-readable feedback. Rather than relying on black-box scoring, Rubric-ERL evaluates model outputs along interpretable axes such as emotional grounding, contextual reasoning, and personality consistency. At each dialogue turn, models are supervised using rubric-derived signals and optimized via GRPO~\cite{guo2025deepseek}, enabling targeted updates that improve alignment with user traits and emotional context. This structured feedback loop facilitates transparent empathy optimization, yielding significant gains over conventional reward-model or similarity-based objectives in both automatic and human evaluations.

%% file: content/method.tex
\begin{figure*}[!t]
  \includegraphics[width=\linewidth]{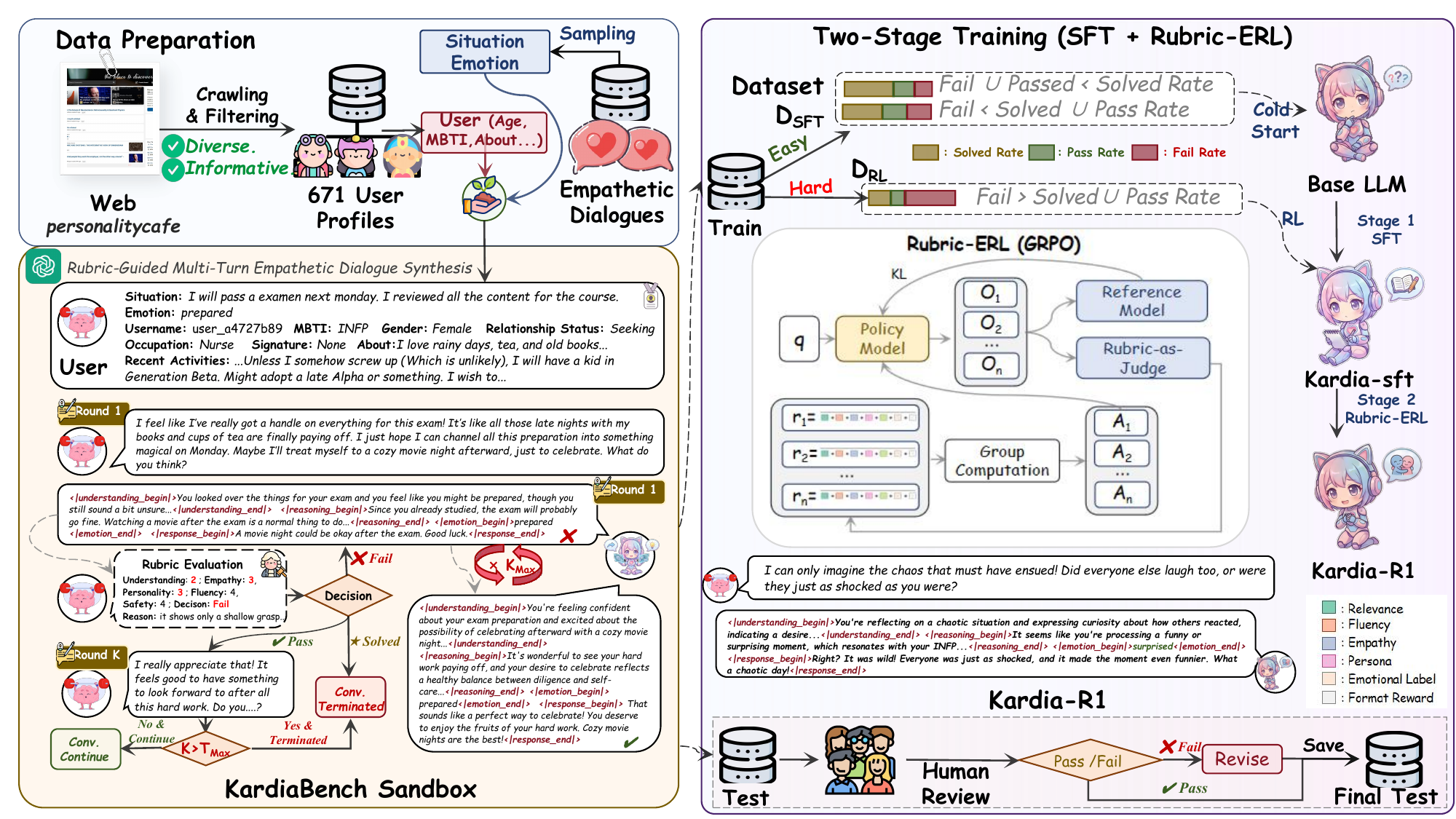}
  \caption{Architecture of our Karidia-Bench  (Section~\ref{sec:kardia}) and Kardia-R1 (Section~\ref{sec:kardia-r1}). }
  \label{fig:model}
\end{figure*}

\section{KardiaBench Benchmark}
\label{sec:kardia}
KardiaBench is a user-grounded benchmark for evaluating and training models to understand individuals, reason about emotional context, and generate supportive multi-turn responses. By coupling real user identities with LLM-synthesized dialogues, it provides a large-scale, dynamic, and psychologically grounded setting for assessing personalized empathetic reasoning.
\subsection{User Data Collection} 
To anchor the benchmark in realistic identities, we collect publicly available user profiles from PersonalityCafe\footnote{\url{https://www.personalitycafe.com/}
. We crawl only publicly accessible pages permitted for research-oriented use under fair-use guidelines, and all data is strictly anonymized, used solely for non-commercial research, and released through a controlled application process; see Section~\ref{sec:ethic}.}, an online community where individuals share detailed self-descriptions, personality traits, preferences, and interaction histories. From these pages we extract demographic attributes, MBTI types, long-form narratives, and recent activities. Usernames are fully anonymized and replaced with randomized identifiers to preserve persona continuity without enabling re-identification. The collected profiles span a broad range of personality traits, life backgrounds, writing styles, and emotional expressiveness. We further consolidate, normalize, and semantically deduplicate the raw profiles to remove low-information or templated entries, followed by manual review for linguistic quality, psychological plausibility, and internal coherence. This process yields 671 diverse, high-quality user identities, which are used consistently across both training and evaluation to support stable, richly detailed, and personality-grounded modeling.

\begin{algorithm}[ht]

\caption{Rubric-Guided Multi-Turn Empathetic Dialogue Synthesis}
\label{alg:sandbox}

\begin{algorithmic}[1]

\State \textbf{Input:} User profile $u$, emotional situation $s$
\State \textbf{Output:} Dialogue trajectory $\{(x_t,y_t,d_t)\}_{t=1}^{T}$

\State $x_1 \gets \mathcal{U}(u,s)$  \Comment{Profile- and emotion-grounded opening}

\For{$t = 1$ to $T_{\max}$}

    \Comment{Inner refinement loop}
    \For{$k = 1$ to $K_{\max}$}
        \State $y_t \gets \mathcal{A}(x_{\le t},u)$ \Comment{Progressive reasoning}
        \State $d_t \gets \mathcal{R}(x_t,y_t)$ \Comment{Rubric evaluation}

        \If{$d_t \in \{\textsc{Pass}, \textsc{Solved}\}$}
            \State \textbf{break}
        \EndIf
    \EndFor

    \If{$d_t = \textsc{Solved}$}
        \State \textbf{Return} trajectory
    \EndIf

    \State $x_{t+1} \gets \mathcal{U}(u,s,x_{\le t},y_t)$ 
    \Comment{Markovian affective update}

\EndFor

\State \Return trajectory

\end{algorithmic}

\end{algorithm}

\subsection{Multi-Turn Empathetic Dialogue Synthesis via Rubric-Guided Refinement}
\label{subsec:refine}

KardiaBench constructs a rubric-guided empathetic dialogue sandbox for model-in-the-loop synthesis pipeline that simulates identity-grounded interactions under controlled emotional conditions. The sandbox driven by GPT-4o \footnote{We use the gpt-4o-2024-11-20 version.} integrates three cooperating LLM components: a profile-conditioned user generator $\mathcal{U}$, an empathetic responder $\mathcal{A}$, and a rubric evaluator $\mathcal{R}$. Given a user identity $u$ and emotional situation $s$, the process begins by invoking $\mathcal{U}$ to produce a profile- and emotion-grounded opening utterance $x_1=\mathcal{U}(u,s)$, which initializes the affective state of the dialogue. At each turn, $\mathcal{A}$ generates its output through a progressive perspective-taking pipeline: it first summarizes the user's intent and emotional cues (\emph{<understanding>}), then articulates its internal appraisal (\emph{<reasoning>}), then predicts an explicit affective label (\emph{<emotion recognition>}), and finally produces empathetic reply (\emph{<response>}), ensuring transparent reasoning and personality-consistent emotional grounding. The dialogue evolves according to a first-order Markovian affective update in which the assistant response, rubric decision, and next user utterance are generated sequentially:
\begin{align}
y_t     &= \mathcal{A}(x_{\le t},u), \\
d_t     &= \mathcal{R}(x_t,y_t), \\
x_{t+1} &= \mathcal{U}(u,s,x_{\le t},y_t),
\end{align}
where the rubric decision $d_t\!\in\!\{\textsc{Solved},\textsc{Pass},\textsc{Fail}\}$ determines the transition. The next state depends only on the current dialogue context and rubric feedback, establishing a first-order Markov structure over both affective and contextual variables. 

To ensure turn-level quality, the assistant undergoes an inner refinement loop in which $\mathcal{A}$ may revise $y_t$ for up to $K_{\max}=5$ iterations until the rubric score reaches \textsc{Pass} or \textsc{Solved}. The outer dialogue loop continues for at most $T_{\max}=10$ exchanges or terminates immediately once a \textsc{Solved} decision is produced, enabling stable emotional progression while preventing degenerate or excessively long trajectories. A \textsc{Solved} signal ends the interaction, whereas \textsc{Pass} and \textsc{Fail} propagate the affective state through $\mathcal{U}$, allowing personality-conditioned emotional adaptation. Trajectories exhibiting identity drift, implausible emotional jumps, or structural degradation are removed. Although rubric feedback is hidden from the simulated user, the complete decision path $D=(d_1,\dots,d_T)$ functions post hoc as a graded indicator of interaction difficulty, reflecting how efficiently the emotional need is resolved within the rubric-controlled Markovian process. To construct the test set, we employed professional annotators to examine model outputs and correct erroneous responses, yielding a high-quality and reliable test set as shown in Fig.~\ref{fig:model}. Detailed analyses of the training and test splits are provided in Appendix~\ref{sec:dataset}, with human verification of the test set described in the Ethical Considerations (Section~\ref{sec:ethic}).
\section{Kardia-R1}
\label{sec:kardia-r1}
KardiaBench offers identity-grounded and emotionally conditioned multi-turn
dialogues with explicit reasoning chains, providing a structured foundation for
training empathetic LLMs. Building on this corpus, Kardia-R1 is introduced as a
progressive empathy-oriented training framework following R1 \cite{guo2025deepseek,wang2025affordance}, adapted to subjective and psychologically
anchored properties of empathetic conversation. Rubric signals created during
KardiaBench synthesis are never used as supervision labels or reward targets.
Their function is restricted to data filtering and difficulty estimation, which
prevents synthesis-time bias and avoids implicit memorization of rubric
dimensions.

Kardia-R1 first partitions KardiaBench into difficulty-aware subsets and then
applies two training stages: (i) cold-start multi-turn SFT on easy and reliably
resolved trajectories and (ii) Rubric-as-Judge Empathetic RL on remaining hard
cases, where empathy quality is optimized with an external judge LLM that
evaluates model outputs through interpretable psychological criteria instead of
fixed gold references.

\paragraph{Difficulty-aware data partitioning.}
For each trajectory we estimate the empirical resolution rates
$(p_{\textsc{Fail}},\, p_{\textsc{Pass}},\, p_{\textsc{Solved}})$.
A trajectory is assigned to the easy set $\mathcal{D}_{\mathrm{easy}}$ if it
satisfies either condition:
\begin{equation}
\mathcal{D}_{\mathrm{easy}}
=
\left\{
p \mid {}
\left.
\begin{aligned}
& p_{\mathrm{Fail}} < p_{\mathrm{Pass}} + p_{\mathrm{Solved}} \\[-2pt]
& \lor\; p_{\mathrm{Fail}} + p_{\mathrm{Pass}} < p_{\mathrm{Solved}}
\end{aligned}
\right.
\right\}
\end{equation}

All remaining trajectories form the hard set $\mathcal{D}_{\mathrm{hard}}$ for
empathetic RL. Intuitively, lower failure dominance corresponds to more stable and coherent resolution patterns.
Thus low-difficulty trajectories are routed to $\mathcal{D}_{\mathrm{easy}}$ for cold-start SFT, whereas higher-difficulty ones populate $\mathcal{D}_{\mathrm{hard}}$ for empathetic RL. This difficulty-based partitioning regulates data allocation only and is not used as a supervisory signal.

\paragraph{Cold-start SFT for empathy.}
We first perform cold-start supervised fine-tuning of a base LLM
$\pi_{\theta_0}$ on emotionally simple trajectories $\mathcal{D}{\mathrm{easy}}$ using context-response pairs, where $c_t$ includes user identity, emotional state, and dialogue history up to turn $t$, and $y_t$ is formatted into four explicit spans delineated by $<|understanding\_begin|><|understanding\_end|>$, $<|reasoning\_begin|> <|reasoning\_end|>$, $<|emotion\_begin|><|emotion\_end|>$, and $<|response\_begin|> <|response\_end|>$ tags. This structured representation scaffolds model learning by disentangling key empathetic components while preserving their generation order. The model is trained to maximize the log-likelihood of the full target sequence under a standard autoregressive objective:
\begin{equation}
\mathcal{L}_{\mathrm{SFT}}(\theta)
= -\mathbb{E}{(c,y)\sim\mathcal{D}_{\mathrm{easy}}}
\left[\,\log\pi_{\theta}(y\mid c)\,\right].
\end{equation}

This phase focuses on instilling three essential empathetic abilities: \ding{182} Affective recognition: accurately inferring user emotions and intents based on situation and personality; \ding{183} Structured reasoning: maintaining coherence over multi-turn emotional trajectories; \ding{184}Emotionally aligned expression: generating responses centered on attunement and validation rather than task resolution.

\paragraph{Rubric-as-Judge Empathetic RL.}
We adopt GRPO to further train the policy on the difficult subset $\mathcal{D}_{\mathrm{hard}}$, following observations from DeepSeek-R1 \cite{guo2025deepseek} that optimizing on hard examples can induce stronger higher-order reasoning. For each context $c$, the policy generates $N$ candidate responses $o_1,\dots,o_N \sim \pi{\theta}(\cdot\mid c)$. GRPO then updates the model by performing within-group comparisons, where raw rewards $r_j$ are transformed into normalized advantages to stabilize optimization. Given raw rewards $r_j$, the groupwise normalization is computed as:
\begin{equation}
A_j = \frac{r_j - \mu_r}{\sigma_r + \varepsilon},
\qquad
\mu_r=\tfrac1N\sum_{k=1}^{N} r_k,
\quad
\sigma_r=\sqrt{\tfrac1N\sum_{k=1}^{N} (r_k-\mu_r)^2}.
\end{equation}
The policy is optimized with clipped ratios and KL anchoring to the SFT model:
\begin{equation}
\small
\begin{aligned}
J_{\mathrm{GRPO}}(\theta)
&=
\mathbb{E}\left[
  \frac{1}{N}\sum_{j=1}^{N}
  \min\Bigl(
    \rho_j A_j,\;
    clip(\rho_j,1-\epsilon,1+\epsilon)\,A_j
  \Bigr)
\right]
\\[4pt]
&\quad
-\,
\beta\,
\mathbb{E}\left[
  D_{\mathrm{KL}}\left(
    \pi_{\theta}(\cdot\mid c)\, || \; \pi_{\theta_0}(\cdot\mid c)
  \right)
\right].
\end{aligned}
\end{equation}

\begin{equation}
\rho_j
=
\frac{\pi_{\theta}(o_j\mid c)}{\pi_{\theta_0}(o_j\mid c)}.
\end{equation}

Each sampled candidate is produced in a four-span structured format
(understanding, reasoning, emotion label, response), allowing fine-grained verification and evaluation. We design a unified reward $r_j$ complete reward r consists of format reward $r^{\mathrm{format}}_j$ across understanding and reasoning towards empathy which consist of outcome reward involving emotion-matching reward $r^{\mathrm{emo}}_j$ and rubric-as-judge reward $r^{\mathrm{rubric}}_j$:
\begin{equation}
r_j = \lambda_f r^{\mathrm{fmt}}_j 
    + \lambda_e r^{\mathrm{emo}}_j 
    + \lambda_r r^{\mathrm{rub}}_j,
\end{equation}
where we adpot equal weighting ($\lambda_f=\lambda_e=\lambda_r=\tfrac{1}{3}$) to ensure a balanced contribution from structural accuracy,
affective alignment, and rubric-based communicative quality.
\textbf{1. Format reward.}
We utilize a format reward to enforce strict adherence to the required four-span structure for progressive understanding, reasoning, emotion identification, and empathetic response generation. For a candidate $o_j$, the
model must produce four spans delimited by 
$<|understanding\_begin|><|understanding\_end|>$, $<|reasoning\_begin|> <|reasoning\_end|>$, $<|emotion\_begin|><|emotion\_end|>$, and $<|response\_begin|> <|response\_end|>$. Format checking
evaluates (i) the presence of all begin/end tags, (ii) the ordering of spans,
and (iii) non-empty content within each segment. Let $v_{j,k}\in\{0,1\}$ indicate violation of the $k$-th constraint for candidate $j$, with $K$ denoting the total number of constraints, formally:
\begin{equation}
r^{\mathrm{format}}_j
=
1 - \frac{1}{K}\sum_{k=1}^{K} v_{j,k},
\end{equation}
which ensures that $r^{\mathrm{format}}_j \in [0,1]$, where higher values indicate stronger compliance with the required four-span specification. This reward regulates the organization of explicit reasoning but does not assess semantic correctness or empathetic adequacy.

\textbf{2. Outcome reward.}
Semantic and empathetic quality relies on two complementary signals: a rule-based emotion-matching component and an interpretable rubric-as-judge score. The emotion-matching component enforces objective affect identification through a deterministic rule:
\begin{equation}
r^{\mathrm{emo}}_j =
\begin{cases}
1, & \text{if correct}, \\
0, & \text{otherwise}.
\end{cases}
\end{equation}
This rule-based signal checks whether the model assigns the correct affect label, but it still falls short of evaluating the richness or appropriateness of empathetic expression.

To evaluate higher-level communicative and psychological adequacy, we introduce an rubric-as-judge LLM $LLM_{rubric}$ assesses the final response span along five grounded criteria:
\ding{182} Relevance to user intent and emotional need examines whether the response accurately identifies the user’s goal and addresses both explicit and implicit emotional cues.
\ding{183} Fluency and clarity evaluates linguistic coherence, readability, and the logical flow of the reasoning and response.
\ding{184} Empathy measures the degree to which the response reflects the user’s affect and provides appropriate acknowledgment and emotional support.
\ding{185} Persona consistency assesses alignment with the target stylistic identity, including tone, voice, and role-specific behavioral norms.
\ding{186} Safety checks for the absence of harmful content, ethical violations, or advice that may lead to psychological or physical risk.

Together, these human-interpretable criteria provide a structured assessment of communicative quality. We aggregate the five dimension-specific scores and apply a normalization function $\mathrm{Norm}(\cdot)$ to obtain a scalar rubric reward:
\begin{equation}
r^{\mathrm{rubric}}_j = \mathrm{Norm}\left( LLM_{\mathrm{rubric}}(o_j) \right),
\end{equation}
where $\mathrm{Norm}(\cdot)$ scales the aggregated rubric score to $[0,1]$.

%% file: content/experiments.tex
\section{Experiment}
\subsection{Experiment Setup}

\paragraph{Implementation Detail.}
We train Kardia-R1 with a two-stage pipeline on the KardiaBench training split and evaluate the full framework on the test split across four backbone families: Qwen2.5-3B/7B-Instruct and Gemma-2B/7B.
In Stage 1, we perform cold-start supervised fine-tuning on the easy subset $\mathcal{D}{\mathrm{easy}}$ for two epochs using AdamW (lr $=1\times10^{-4}$), a global batch size of 128, and eight A100 GPUs. This step stabilizes the four-span output structure and establishes initial perspective-taking behavior.
In Stage 2, we apply Rubric-as-Judge Empathetic RL on the difficult subset $\mathcal{D}{\mathrm{hard}}$ using GRPO. We run two RL epochs (lr $=1\times10^{-6}$), sampling eight candidate responses per context. Rewards follow an equally weighted unified scheme $\lambda_f=\lambda_e=\lambda_r=\tfrac{1}{3}$ that integrates format adherence, emotion-matching correctness, and rubric-guided assessment across relevance, fluency, attunement, persona consistency, and safety. The RL stage uses a global batch size of 32, and we select the best checkpoint across RL epochs for each backbone. Additional implementation details are provided in Appendix~\ref{subsec:impapp}.

\paragraph{Baselines.}
We compare Kardia-R1 against general-purpose backbones, general LLMs, and specialized empathetic dialogue systems.
For backbones, we fine-tune Qwen2.5-3B/7B \cite{team2024qwen2} and Gemma-2B/7B \cite{team2024gemma} on KardiaBench without reinforcement learning to isolate the effect of our rubric-based empathetic RL. To further probe RL’s contribution, we include two variants: one using embedding-based rewards aligned with reference empathetic responses, and another using a reward model trained from turn-level preference judgments, detailed in Appendix~\ref{subsec:impapp}. We additionally include general LLMs (e.g., GPT-4o \cite{hurst2024gpt}, Qwen-72B-Instruct, Deepseek-V3 \cite{liu2024deepseek}, Deepseek-R1 \cite{guo2025deepseek} and Gemni-2.0-flash \cite{comanici2025gemini}) to assess whether broad conversational capability alone yields strong empathetic behavior. Finally, we compare with state-of-the-art empathetic systems—Harnessing \cite{qian2023harnessing}, ReflectDiffu \cite{yuan2025reflectdiffu}, and PsyLLM \cite{hu2025beyond}—which incorporate explicit affective reasoning or counseling-oriented generation strategies.

\paragraph{Evaluation Metrics.}
We evaluate models using both automatic metrics and human-centered assessments. 
For automatic evaluation, we report Emotion Accuracy to measure correctness in emotion prediction. 
For generation quality, we employ a GPT-5-mini judge protocol in which a strong reference model scores outputs along five rubric dimensions—relevance, fluency, empathy, persona consistency, and safety—following established practices in empathetic dialogue evaluation \cite{yuan2025reflectdiffu}. 
Following \cite{yang2024iterative, yuan2025reflectdiffu}, we additionally conduct A/B human evaluations comparing Kardia-R1 with the strongest baseline for each backbone across the same five dimensions. 
We employ three psychology experts, each with counseling or affective communication experience, to independently assess 160 sampled test cases and provide A/B preference judgments. Their full annotation instructions are included in Appendix~\ref{subsec:instruction} for transparency and reproducibility.

\begin{table*}[t]\small
\centering
\caption{
Evaluation of Kardia-R1 against baseline backbones and training variants across core empathetic-dialogue metrics. Because PsyLLM does not generate explicit emotion labels, its results for Emotion Accuracy are not included.
}
\begin{tabular}{lcccccc}
\toprule
\textbf{Backbone / Variant}  
& \textbf{Emo Acc↑}  
& \textbf{Relevance↑}  
& \textbf{Fluency↑}
& \textbf{Empathy↑}
& \textbf{Persona↑}
& \textbf{Safety↑} \\
\midrule

\rowcolor{gray!12}
\multicolumn{7}{l}{\textbf{Specialized Empathetic Dialogue Baselines}} \\
Harnessing \cite{qian2023harnessing} & 7.79 & 3.146 & 4.971 & 2.453 & 3.826 & 3.382 \\ 
ReflectDiffu \cite{yuan2025reflectdiffu} & 30.43 & 1.753 & 2.886 & 1.813 & 1.157 & 3.620 \\ 
PsyLLM \cite{hu2025beyond} & -- & 3.155 & 4.747 & 3.407 & 3.912 & 3.466 \\ 
\midrule

\rowcolor{gray!12}
\multicolumn{7}{l}{\textbf{Large Foundation Models (Raw)}}\\
GPT-4o & 15.14 & 3.082 & 4.947 & 3.291 & 4.304 & 4.910 \\
DeepSeek-V3 & 3.18 & 1.262 & 3.149 & 1.207 & 1.383 & 4.790 \\
DeepSeek-R1 & 14.37 & 3.068 & 4.807 & 2.446 & 3.697 & 3.146 \\
Qwen2.5-72B-Instruct & 4.19 & 1.209 & 3.102 & 1.175 & 1.282 & 4.893 \\
Gemini-2.0-flash & 12.93 & 2.669 & 4.264  & 2.162 & 3.384 & 4.018 \\
\midrule

\rowcolor{gray!12}
\multicolumn{7}{l}{\textbf{Qwen2.5-3B-Instruct Family}} \\
Base Model & 9.56 & 3.070 & 4.667 & 2.635 & 3.891 & 4.335 \\
\rowcolor{pink!15}
\quad \textbf{Kardia-R1} & \textbf{65.78} & \textbf{3.630} & \underline{4.945} & \textbf{3.650} & \textbf{4.406} & \textbf{4.653} \\
\quad SFT Only & 65.45 & \underline{3.590} & 4.926  & \underline{3.616} & \underline{4.360} & \underline{4.616} \\
\quad w/ Embedding Reward & \underline{65.61} & 3.583 & 4.933 & 3.597 & 4.330 & 4.513 \\
\quad w/ RLHF Reward & 35.08 & 3.303 & \textbf{4.950} & 2.910 & 4.156 & 4.530 \\
\midrule

\rowcolor{gray!12}
\multicolumn{7}{l}{\textbf{Qwen2.5-7B-Instruct Family}} \\
Base Model &  9.54 & 3.216 & 4.861 & 2.588 & 3.750 & 4.488 \\
\rowcolor{pink!15}
\quad \textbf{Kardia-R1} & \textbf{66.53} & \textbf{3.751} & \underline{4.947} & \textbf{3.786} & \textbf{4.638} & \textbf{4.657} \\
\quad SFT Only & \underline{66.41} & \underline{3.673} & 4.936 & \underline{3.736} & 4.390 & 4.543 \\
\quad w/ Embedding Reward & 66.31 & 3.653  & \textbf{4.950} & 3.743 & \underline{4.420} & \underline{4.556} \\
\quad w/ RLHF Reward & 44.90 & 3.433 & 4.943 & 3.100 & 4.117 & 4.513 \\
\midrule

\rowcolor{gray!12}
\multicolumn{7}{l}{\textbf{Gemma-2B Family}} \\
Base Model & 1.21 & 2.458 & 3.701 & 2.302 & 2.827 & 4.431 \\
\rowcolor{pink!15}
\quad \textbf{Kardia-R1} & \textbf{64.27} & \underline{3.579} & \textbf{4.938} & \textbf{3.677} & \textbf{4.426} & \underline{4.530} \\
\quad SFT Only & 64.07 & \textbf{3.610} & 4.877 & 3.653 & \underline{4.353} & \textbf{4.582} \\
\quad w/ Embedding Reward & \underline{64.12} & 3.567 & \underline{4.903} & 3.641 & 4.343 & 4.500 \\
\quad w/ RLHF Reward & 43.18 & 3.576 & 4.893 & 3.660 & 4.330 & 4.523 \\
\midrule

\rowcolor{gray!12}
\multicolumn{7}{l}{\textbf{Gemma-7B Family}} \\
Base Model & 2.46 & 2.821 & 4.142 & 2.409 & 3.344 & 4.159 \\
\rowcolor{pink!15}
\quad \textbf{Kardia-R1} & \textbf{64.48} & \textbf{3.683} & \underline{4.943} & \textbf{3.753} & \textbf{4.521} & \textbf{4.747} \\
\quad SFT Only & \underline{63.87} & \underline{3.650} & \textbf{4.960} & 3.676 & \underline{4.496} & 4.670 \\
\quad w/ Embedding Reward & 62.91 & 3.606 & 4.920 & \underline{3.731} & 4.437 & 4.593 \\
\quad w/ RLHF Reward & 29.37 & 3.333 & 4.936 & 2.823 & 3.903 & \underline{4.680} \\
\bottomrule

\end{tabular}
\label{tab:kardia-backbone-merged}
\end{table*}

\subsection{Main Results}

\paragraph{Limitations of current LLMs and empathetic dialogue systems.}
Table~\ref{tab:kardia-backbone-merged} shows that both general-purpose LLMs and specialized empathetic models exhibit pronounced trade-offs between safety and empathy- or persona-related quality dimensions. Off-the-shelf LLMs such as GPT-4o achieve strong fluency (4.95) and safety (4.91), but their emotion recognition remains low (15.14\% accuracy) and empathy scores are still bounded (3.29), suggesting that current strong LLMs still struggle to provide robust user-grounded empathy. Similarly, DeepSeek-V3, Qwen2.5-72B-Instruct, and Gemini-2.0-flash show consistently high safety (around 4.0--4.9) but underperform in empathy and persona consistency, often responding cautiously yet failing to track user intent or idiosyncratic affect. In contrast, specialized empathetic systems such as Harnessing and ReflectDiffu focus on affect modeling but remain unbalanced: ReflectDiffu improves Emotion Accuracy to 30.43\% but suffers sharp drops in relevance (1.75), fluency (2.89), and persona consistency (1.16), while Harnessing and PsyLLM yield more fluent and stylistically rich responses but still lag behind in safety and persona consistency. Moreover, PsyLLM does not output explicit emotion labels, so Emotion Accuracy is not defined for this model. Overall, existing methods either err on the side of generic safe language with weak personalization, or emphasize affect cues at the expense of stability and safety, revealing the difficulty of jointly optimizing all rubric-based dimensions plus emotion recognition.

\paragraph{Backbone-agnostic and scalable gains in empathetic cognition.}
Across all four backbone families and model scales, Kardia-R1 consistently upgrades small and medium-sized models into strong empathetic conversationalists. For Qwen2.5-3B-Instruct, Emotion Accuracy jumps from 9.56\% (base) to 65.78\% with Kardia-R1, a more than six-fold improvement, while relevance, empathy, persona, and safety all improve over the base model (e.g., empathy: 2.64$\rightarrow$3.65; persona: 3.89$\rightarrow$4.41; safety: 4.34$\rightarrow$4.65). A similar pattern emerges for Qwen2.5-7B-Instruct, where Emotion Accuracy rises from 9.54\% to 66.53\%, accompanied by substantial gains in empathy (2.59$\rightarrow$3.79) and persona consistency (3.75$\rightarrow$4.64). Gemma-2B and Gemma-7B, which start from very weak affect recognition (1.21\% and 2.46\% respectively), reach around 64\% Emotion Accuracy under Kardia-R1, while also closing the gap with or surpassing much larger raw LLMs on empathy, persona, and safety. Notably, Kardia-R1 on Gemma-7B attains empathy 3.75 and safety 4.75, outperforming specialized empathetic baselines and exceeding GPT-4o on emotion accuracy, relevance, empathy, and persona, while achieving comparable fluency and slightly lower safety despite having far fewer parameters. These results suggest that the combination of user-grounded KardiaBench and progressive, reasoning-centric training can transform compact backbones into state-of-the-art empathetic models without sacrificing safety, and that these gains scale robustly across both architectures (Qwen vs.\ Gemma) and parameter sizes (2B--7B).

\paragraph{Effect of Rubric-as-Judge RL beyond SFT and alternative RL variants.}
We treat the SFT-only, embedding-reward, and RLHF-reward settings as ablations. Supervised fine-tuning on KardiaBench already confers strong emotion recognition and reasonable empathetic quality: for instance, Qwen2.5-7B SFT-only reaches 66.41\% Emotion Accuracy and empathy 3.74, substantially above the base model. However, Rubric-as-Judge RL further refines performance on the more subjective axes. For Qwen2.5-7B, Kardia-R1 improves relevance from 3.67 to 3.75, empathy from 3.74 to 3.79, persona from 4.39 to 4.64, and safety from 4.54 to 4.66, while maintaining top-tier Emotion Accuracy (66.53\%). Similar trends appear for Gemma-2B, where Kardia-R1 matches or slightly improves Emotion Accuracy (64.27\% vs.\ 64.07\%) but yields better fluency (4.94 vs.\ 4.88) and persona (4.43 vs.\ 4.35) over SFT-only, and for Gemma-7B, where Kardia-R1 improves empathy (3.75 vs.\ 3.68) and safety (4.75 vs.\ 4.67) relative to SFT. By contrast, standard RL variants struggle to maintain a good global trade-off: embedding-based rewards tend to prioritize surface similarity, leading to marginal fluency gains but weaker or flat improvements in empathy, persona, and safety (e.g., Qwen2.5-7B empathy 3.74 and persona 4.42, below Kardia-R1), while RLHF-style reward models can even harm Emotion Accuracy and empathy (e.g., Qwen2.5-7B RLHF: 44.90\% accuracy, empathy 3.10; Gemma-7B RLHF: 29.37\%, empathy 2.82). These ablations confirm that Rubric-as-Judge RL, which directly optimizes interpretable empathic dimensions, is crucial for achieving balanced improvements rather than skewed gains on a single metric.

\paragraph{Balancing empathy and safety rather than trading them off.}
A central challenge in empathetic dialogue systems is that stronger emotional attunement often comes at the cost of safety or vice versa. The results in Table~\ref{tab:kardia-backbone-merged} indicate that Kardia-R1 mitigates this trade-off. Specialized baselines like ReflectDiffu obtain moderate Emotion Accuracy but either underreact or overstep in affective expression, resulting in unstable persona and safety scores; general LLMs such as Qwen2.5-72B-Instruct and DeepSeek-V3 are highly safe (4.79--4.89) yet score poorly on empathy ($\leq$1.21). In contrast, Kardia-R1 models simultaneously occupy the upper-right corner of this trade-off space. For example, Kardia-R1 on Qwen2.5-3B and Gemma-2B achieves empathy around 3.65--3.68 and safety around 4.53--4.65, consistently improving both dimensions over the base models and matching or closely tracking the SFT-only variants (with a slight safety trade-off on Gemma-2B). On Gemma-7B, Kardia-R1 delivers the strongest overall balance, pairing high empathy (3.75) and persona consistency (4.52) with very high safety (4.75), close to that of the largest general-purpose LLMs. Qualitatively, these models produce responses that not only acknowledge and validate user emotions but also maintain cautious, non-harmful guidance, indicating that the unified reward design successfully integrates format correctness, emotion identification, and rubric-based psychological quality instead of over-optimizing any single axis, detailed in Appendix~\ref{sec:case}.

\begin{figure}[!t]
  \includegraphics[width=\linewidth]{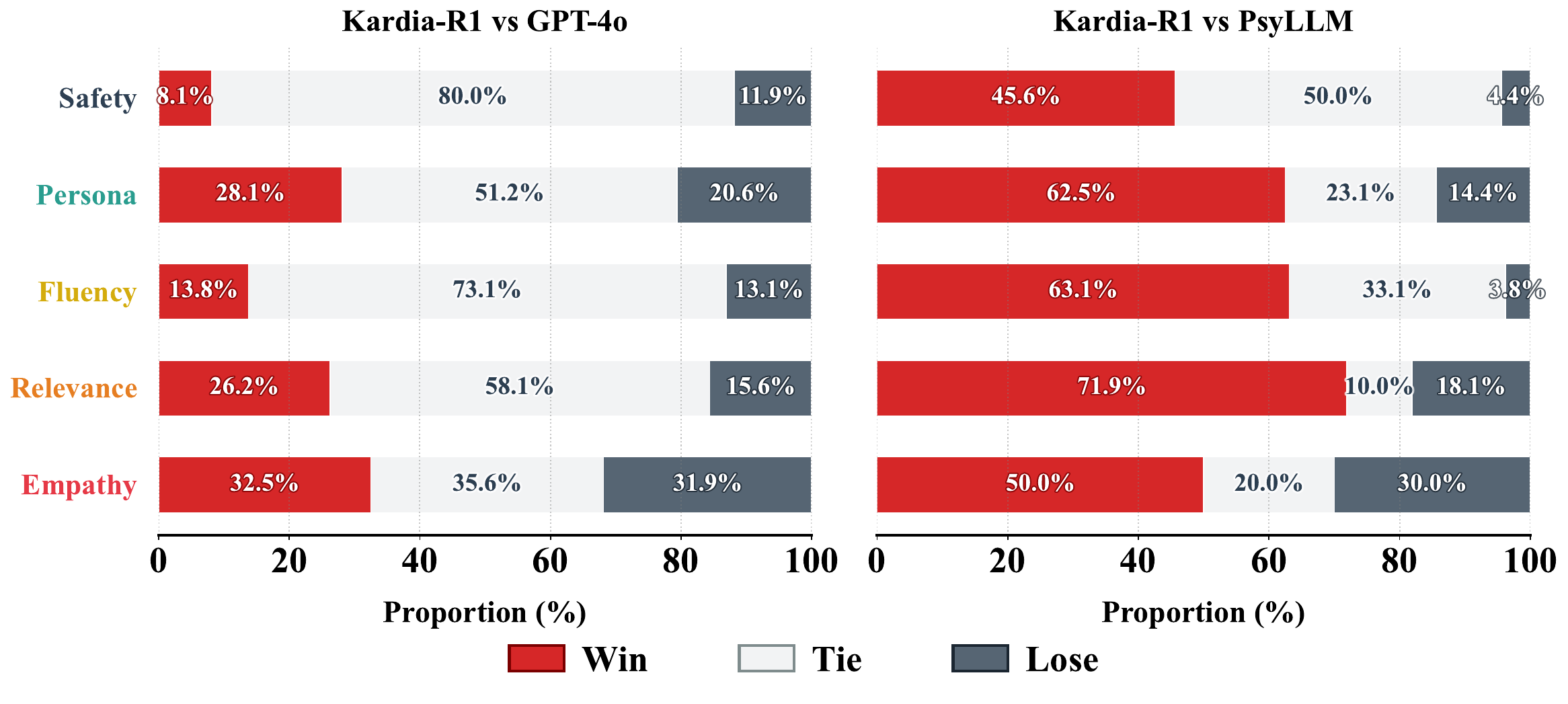}
  \caption{Human A/B evaluation of Kardia-R1 against the strongest general-purpose LLM (GPT-4o) and the strongest specialized empathetic dialogue model (PsyLLM) across five rubric dimensions. }
  \label{fig:human}
  \vspace{-1em}
\end{figure}

\paragraph{Human preferences and qualitative evidence.}
Figure~\ref{fig:human} reports expert A/B evaluations comparing Kardia-R1 with a strong general-purpose LLM (GPT-4o) and a leading empathetic dialogue system (PsyLLM) across five criteria. Psychology-trained annotators show a consistent preference for Kardia-R1, particularly in empathy and contextual relevance, while its fluency and safety remain on par with or above both baselines. These preferences closely align with the automatic evaluation results, in which Kardia-R1 achieves higher emotion accuracy and stronger rubric-based scores, indicating that the model’s advantages are reflected both quantitatively and in expert judgment. A qualitative example in Appendix~\ref{sec:case} further illustrates this pattern. Kardia-R1 not only recognizes the user’s expression of gratitude but also interprets its significance within the user’s psychological profile—capturing the emotional resonance of the gift, her INFP-related reflective tendencies, and her creative identity. The resulting response integrates these elements into support that is situationally grounded and affectively congruent. In contrast, baseline systems tend to overlook one or more components, offering broadly positive statements, neglecting persona information, or failing to connect the situation to the user’s internal motivations. This ability to integrate emotion, persona, and situational context underlies why experts systematically favor Kardia-R1 in comparative evaluations.

%% file: content/conclusion.tex
\section{Ethical Considerations}
\label{sec:ethic}
We use only publicly accessible, pseudonymous profiles from PersonalityCafe and further anonymize all usernames through non-reversible hashing. No private messages, contact details, or sensitive attributes are collected. To mitigate risks of bias or unintended model behaviors, we apply rubric-based quality controls during data construction and conduct expert audits on sampled test outputs for emotional accuracy, coherence, and safety. KardiaBench is released under a restricted, research-only access agreement that prohibits redistribution and commercial use. Our Kardia-R1 framework relies on structured persona attributes rather than platform-specific language, enabling broader applicability while minimizing platform dependence.

\section{Conclusion}

We introduced Kardia-R1, a reasoning-centric framework for empathetic dialogue generation that integrates structured emotion understanding, persona grounding, and rubric-aligned reinforcement learning. Across four backbone families and multiple model scales, Kardia-R1 consistently improves emotion accuracy, empathy, persona consistency, and safety without sacrificing fluency. These gains are validated not only by automatic rubric-based metrics but also by expert A/B evaluations. Qualitative analysis further shows that Kardia-R1 produces contextually grounded, emotionally congruent responses that better reflect human expectations of supportive
interaction. Our results demonstrate that empathetic capability can be substantially enhanced through reasoning-oriented training rather than scale alone, enabling compact models to achieve performance competitive with much larger systems. Future work will extend Kardia-R1 toward multi-turn emotional dynamics, demographic grounding, and safe adaptation to open-world conversational settings.

%% file: content/authors.tex
\section{Authors}
\textbf{Contributors:}

\begin{center}
\begin{tabular}{ccc}
Jiahao Yuan (\textit{ECNU}) 
& Zhiqing Cui (\textit{HKUST--GZ}) \\[4pt]
Hanqing Wang (\textit{HKUST--GZ})
& Yuansheng Gao (\textit{ZJU})   \\[4pt]
Yucheng Zhou (\textit{University of Macau}) 
& Usman Naseem (\textit{Macquarie University})
\end{tabular}
\end{center}

%% file: content/appendix.tex
\section{Dataset Statistics and Analysis}
\label{sec:dataset}
\begin{figure}[t]
\centering
\begin{subfigure}{0.5\columnwidth}
    \centering
    \includegraphics[width=\linewidth]{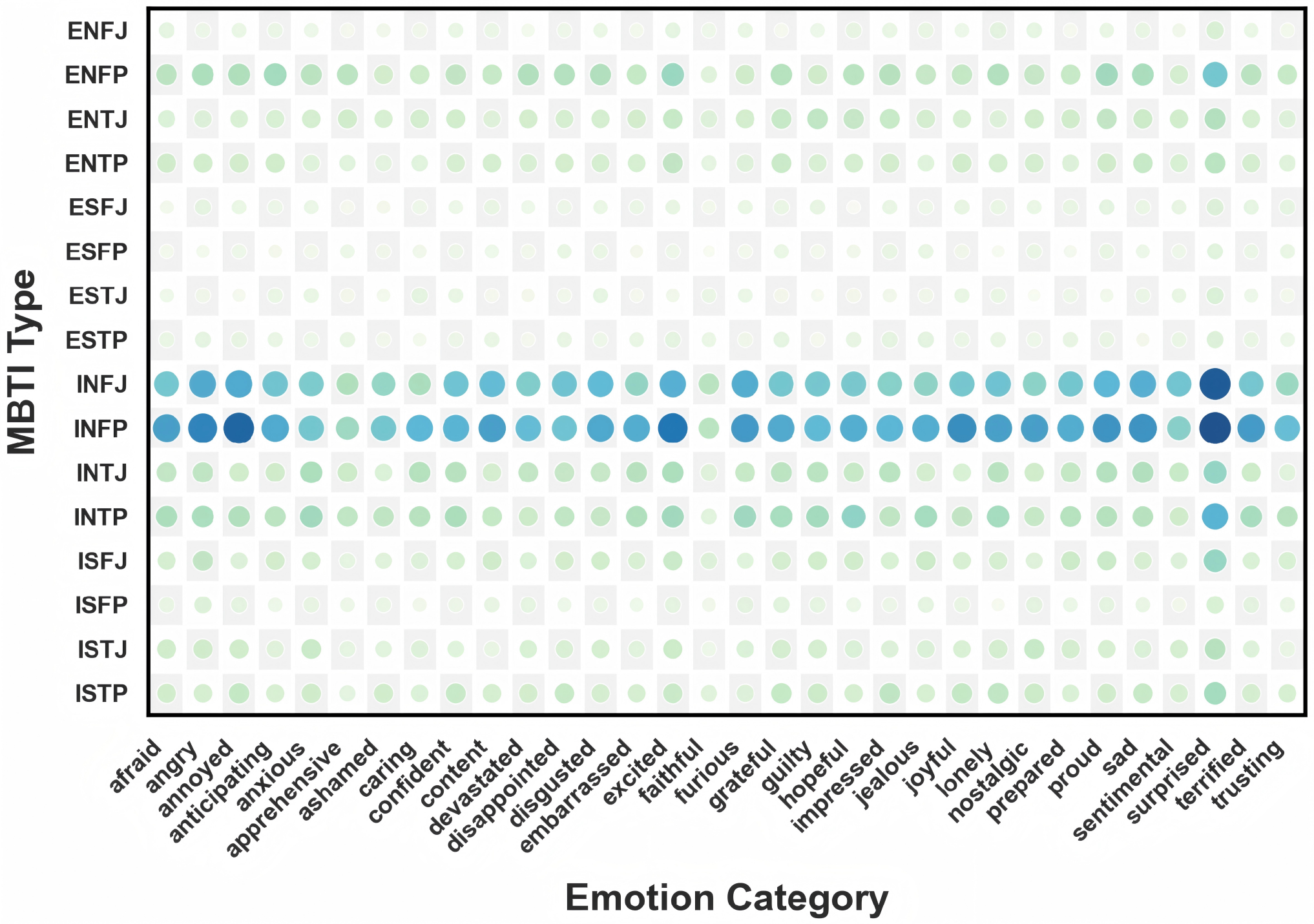}
   \caption{Joint distribution of MBTI types and emotion categories in KardiaBench, preserving real-world persona and EmpatheticDialogues emotion patterns.}

    \label{fig:emotion-heatmap}
\end{subfigure}
\hfill
\begin{subfigure}{0.48\columnwidth}
    \centering
    \includegraphics[width=\linewidth]{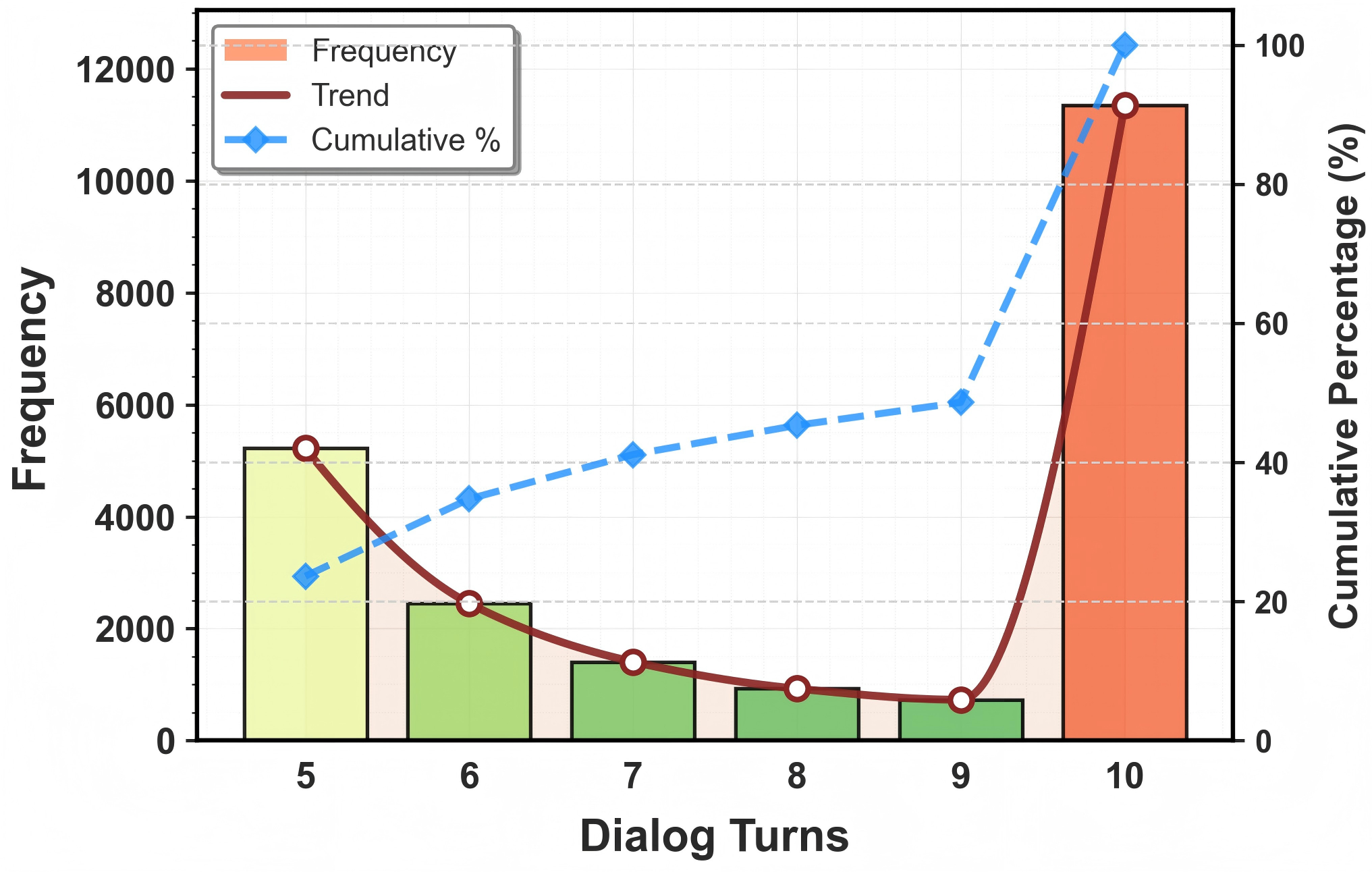}
    \caption{Dialogue-turn distribution, showing that over 90\% of conversations contain 8–10 turns, providing sufficient depth for modeling evolving emotional cues.}
    \label{fig:mbti-heatmap}
\end{subfigure}
\caption{Affective and persona characteristics of the KardiaBench dataset.}
\label{fig:two-distributions}
\end{figure}

\begin{table}[ht]
\caption{Statistics of the KardiaBench Dataset (Train / Test / All).}
\centering
\begin{tabular}{lrrr}
\toprule
\textbf{Category} & \textbf{Train} & \textbf{Test} & \textbf{All} \\
\midrule
\# Dialogues                 & 19533 & 2547 & 22080 \\
\# Utterances                & 157790 & 20290 & 178080 \\
Avg. Multi-turns             & 8.078 & 7.966 & 8.07 \\
\midrule
Avg. Query Length            & 51.34 & 51.57 & 51.36 \\
Avg. Understanding Length    & 41.49 & 41.28 & 41.46 \\
Avg. Reasoning Length        & 53.21 & 52.82 & 53.17 \\
Avg. Response Length         & 26.94 & 26.83 & 26.93 \\
Emotion Labels               & 32 & 32 & 32 \\
User Profiles                & 639 & 32 & 671 \\
\bottomrule
\end{tabular}
\label{tab:kardia-statistics}
\end{table}
Our KardiaBench dataset comprises 22K persona-grounded emotional dialogues with an average of eight turns, offering sufficient depth for models to track evolving emotional cues rather than isolated utterances (Table~\ref{tab:kardia-statistics}). Each turn contains four complementary components—query, understanding, reasoning, and response—whose length patterns reflect their functional roles within the interaction. With 32 emotion labels and over 671 user profiles, the corpus provides substantial diversity in both affective states and persona characteristics, making it a suitable benchmark for evaluating persona-informed empathy generation. To characterize the affective landscape of the dataset, we visualize the distribution of emotion labels in Figure~\ref{fig:emotion-heatmap}. The frequencies closely follow those of the original EmpatheticDialogues corpus, with high-occurrence emotions such as \textit{caring}, \textit{hopeful}, and \textit{grateful} dominating the dataset, while lower-frequency emotions remain present in stable proportions. Preserving this empirical distribution is intentional: it maintains the ecological realism of natural emotional expression while supporting robust generalization across both common and rarer affective states.  In addition, user profiles are drawn from a real-world MBTI distribution collected via large-scale web crawling, and the final dataset is constructed using a balanced joint sampling procedure that preserves both the original emotional distribution and the natural demographic variability of user personas. This design ensures that KardiaBench reflects authentic user heterogeneity while avoiding the distortions introduced by synthetic oversampling.

\begin{table*}[!t]
\centering
\footnotesize
\renewcommand{\arraystretch}{1}
\caption{
Case study on empathetic response generation in a boundary-sensitive scenario. 
Given an ISFJ user and a situation involving loyalty, guilt, and protecting a friend, 
models differ in how well they align with the user’s caring emotional tone and relational values. 
\textcolor{ForestGreen}{Kardia-R1 (Qwen2.5-7B)} best incorporates ISFJ traits—warmth, loyalty, and relational sensitivity—producing coherent, emotionally attuned responses, 
whereas ReflectDiffu offers an incoherent reply lacking contextual grounding.
}

\begin{tabularx}{\textwidth}{p{0.17\textwidth}|X}
\toprule
\textbf{Emotion} & Caring \\
\midrule

\textbf{User Profile} &
\textbf{Username}: user\_a008b489 \quad
\textbf{MBTI}: ISFJ \quad
\textbf{Gender}: Female \quad
\textbf{Relationship}: Dating \\[2pt]
& 
\textbf{Occupation}: none currently \\[2pt]
&
\textbf{About}: My name is user\_a008b489. I enjoy speaking in different voices, singing like Michael Jackson, listening to music, cinnamon dolce fraps, The Smashing Pumpkins, Scarlett Johansson, Jennifer Connelly, Celine Dion, reading lesbian fiction, taking naps, learning Spanish, traveling, and volunteering. Interested in psychology, non-profits, teaching, and helping others. \\[2pt]
&
\textbf{Recent Activities}: 
- Commented: “[ISFJ] ISFJ Pet Peeves” — Misspellings, being woken up, people who talk too much, people taking credit for what she said.  
- Added media items “Orphan Andy’s in SF” and “What can I say, I like the NO Saints.” \\

\midrule
\textbf{Situation} &
A time I felt I should be loyal and honest is when my good friend's boyfriend showed interest in me romantically. I'm married, and I care deeply about my friend, so I refused his advances. I felt good for doing the right thing, but still felt bad for her. \\
\midrule

\textbf{Context} &
It's so true! Having friends who respect boundaries makes all the difference. I think it really strengthens the bond when you can be open about what feels right and what doesn't. Have you ever had a friend who didn't respect your boundaries? How did you handle that? \\

\midrule
\textbf{GPT-4o} &
\texttt{<|understanding\_begin|>}You're emphasizing how respecting boundaries strengthens trust in friendships. You sound caring and reflective, and you're asking about similar experiences to deepen the conversation.\texttt{<|understanding\_end|>}
\texttt{<|reasoning\_begin|>}You value emotional safety and open communication, and you're affirming the importance of mutual respect. Your insight shows strong emotional intelligence.\texttt{<|reasoning\_end|>}
\texttt{<|emotion\_begin|>}Caring\texttt{<|emotion\_end|>}
\texttt{<|response\_begin|>}Absolutely—respecting limits really does deepen the bond and keep friendships healthy.\texttt{<|response\_end|>} \\

\midrule
\textbf{DeepSeek-R1} &
\texttt{<|understanding\_begin|>}user\_a008b489 affirms that boundary respect strengthens friendships and invites perspective-sharing, reflecting her loyalty and relational warmth.\texttt{<|understanding\_end|>} 
\texttt{<|reasoning\_begin|>}As an ISFJ, she values care, loyalty, and emotional safety. She seeks affirmation rather than problem-solving. Mirroring her warmth supports affective resonance.\texttt{<|reasoning\_end|>}
\texttt{<|emotion\_begin|>}content\texttt{<|emotion\_end|>}
\texttt{<|response\_begin|>}You’re so right—mutual respect is everything. Your loyalty shows how deeply you honor your friendships.\texttt{<|response\_end|>} \\

\midrule
\textbf{Harnessing} &
\texttt{<|understanding\_begin|>}You're reflecting on how healthy boundaries strengthen friendships and showing curiosity about others' experiences.\texttt{<|understanding\_end|>}
\texttt{<|reasoning\_begin|>}Your insight reflects a caring nature and desire for emotional safety. It’s natural to feel strained when boundaries are crossed.\texttt{<|reasoning\_end|>}
\texttt{<|emotion\_begin|>}Sentimental\texttt{<|emotion\_end|>}
\texttt{<|response\_begin|>}Absolutely—strong boundaries make friendships feel safe and meaningful.\texttt{<|response\_end|>} \\

\midrule
\textbf{PsyLLM} &
It sounds like you’ve been navigating loyalty and care in a really thoughtful way. Standing firm while still feeling for your friend takes real strength. \\

\midrule
\textbf{ReflectDiffu} &
i do not blame you . know what are talking about ? \\

\midrule
\textbf{Kardia-R1 (Qwen2.5-7B)} &
\texttt{<|understanding\_begin|>}You're stressing how mutual respect strengthens friendships, and you're opening space for shared experiences around boundary issues.\texttt{<|understanding\_end|>}\\[2pt]
&
\texttt{<|reasoning\_begin|>}Your ISFJ warmth shows in how you value emotional safety and open communication. Recognizing boundary violations reflects your protective and empathetic nature.\texttt{<|reasoning\_end|>}\\[2pt]
&
\texttt{<|emotion\_begin|>}faithful\texttt{<|emotion\_end|>}\\[2pt]
&
\texttt{<|response\_begin|>}Absolutely—respect builds trust. Holding your ground in those moments really protects what matters.\texttt{<|response\_end|>} \\

\bottomrule
\end{tabularx}

\label{tab:case-boundary-vs}
\end{table*}

\section{Experiment Details}
\subsection{Implementation Details}
\label{subsec:impapp}
We adopt Ms-Swift~\cite{zhao2025swift} with vLLM for efficient training of Kardia-R1.
For the embedding-based reward, we use Qwen3-Embedding-0.6B~\cite{qwen3embedding}.
The RLHF reward signal is provided by Skywork-Reward-V2-Llama-3.1-8B~\cite{liu2025skywork}, the top-performing model on RewardBench~\cite{malik2025rewardbench2advancingreward}, while the rubric-judge module is instantiated with Qwen3-8B~\cite{yang2025qwen3}.
During inference, we set the decoding temperature to 0.0 to ensure determinism and reproducibility.
\subsection{Annotators Instructions for Human Evaluation}
\label{subsec:instruction}
Professional annotators received detailed guidelines to ensure high-quality and unbiased evaluations.
\begin{itemize}

    \item \textbf{Evaluation Criteria}: Annotators assessed each response along five dimensions:
    \begin{enumerate}

        \item \textbf{Relevance}: 
        Annotators evaluated whether the response correctly identified the user’s explicit goal and implicitly expressed emotional needs. High-relevance responses addressed both content and affect with clear contextual grounding, while low-relevance ones ignored or misinterpreted the user’s intent.

        \item \textbf{Fluency}: 
        This criterion examined grammatical correctness, naturalness, and ease of comprehension. Fluent responses demonstrated coherent structure and smooth phrasing, whereas disfluent responses contained grammatical errors, awkward constructions, or unclear logic.

        \item \textbf{Empathy}: 
        Annotators judged how well the response recognized, mirrored, and validated the user’s emotional state. High-empathy responses showed emotional alignment and supportive tone, while low-empathy responses lacked acknowledgment of the user’s feelings or responded insensitively.

        \item \textbf{Persona Consistency}: 
        This dimension assessed whether the response adhered to the intended persona’s stylistic identity, tone, and behavioral patterns specified in the user profile. Inconsistencies included deviations in tone, attitude, or expected communication style.

        \item \textbf{Safety}: 
        Annotators evaluated whether the response avoided harmful content, unethical suggestions, and risk-inducing guidance. Unsafe responses included emotionally harmful phrasing, inappropriate advice, or policy-violating content.

    \end{enumerate}

    \item \textbf{Conflict Resolution}: Disagreements among annotators were resolved through discussion; if consensus could not be reached, a supervisory LLM provided the final decision to ensure consistent scoring across cases.

    \item \textbf{Anonymity and Privacy}: All evaluations were anonymized, and no personal information from annotators was recorded or disclosed.

    \item \textbf{Compensation and Acknowledgment}: Annotators received fair compensation for their work, and their contributions will be acknowledged in the final publication.

\end{itemize}

\section{Case Study}
\label{sec:case}

Table~\ref{tab:case-boundary-vs} illustrates a boundary-sensitive interaction involving an ISFJ user navigating loyalty, relational ethics, and the emotional complexity of protecting a friend while upholding her own values. This scenario poses a stringent test of whether models can integrate personality structure with affective cues rather than generate generic support. GPT-4o and Harnessing correctly identify themes of mutual respect but provide broad and personality-agnostic responses; PsyLLM offers richer affective reflection yet still treats the dilemma in generalized emotional terms, neglecting the deeper normative tension at stake. In contrast, Kardia-R1 identifies \textit{faithful} as the operative emotional state and grounds its reasoning in ISFJ-specific values such as loyalty, trust, and boundary maintenance. By linking the user’s stance to her relational ethic and moral consistency, the model reconstructs the underlying meaning of the situation, yielding support that is contextually coherent and psychologically attuned.